\documentclass[conference]{IEEEtran}
\IEEEoverridecommandlockouts
\usepackage{cite}
\usepackage{amsmath,amssymb,amsfonts}
\usepackage{algorithmic}
\usepackage{graphicx}
\usepackage{textcomp}
\usepackage{xcolor}
\begin{document}

\title{Improving Intersession Reproducibility for Forearm Ultrasound based Hand Gesture Classification through an Incremental Learning Approach\\
\thanks{*Co-first authors}
}

\author{\IEEEauthorblockN{Keshav Bimbraw*}
\IEEEauthorblockA{\textit{Robotics Engineering} \\
\textit{Worcester Polytechnic Institute}\\
Worcester, USA \\
kbimbraw@wpi.edu}
\and
\IEEEauthorblockN{Jack Rothenberg*}
\IEEEauthorblockA{\textit{Robotics and Biomedical Engineering} \\
\textit{Worcester Polytechnic Institute}\\
Worcester, USA \\
jarothenberg@wpi.edu}
\and
% \IEEEauthorblockN{Keshav Bimbraw*}
% \IEEEauthorblockA{\textit{Robotics Engineering} \\
% \textit{Worcester Polytechnic Institute}\\
% Worcester, USA \\
% kbimbraw@wpi.edu}
% \and
\IEEEauthorblockN{Haichong K. Zhang}
\IEEEauthorblockA{\textit{Robotics and Biomedical Engineering} \\
\textit{Worcester Polytechnic Institute}\\
Worcester, USA \\
hzhang10@wpi.edu}
}

\maketitle

\begin{abstract}
Ultrasound images of the forearm can be used to classify hand gestures towards developing human machine interfaces. In our previous work, we have demonstrated gesture classification using ultrasound on a single subject without removing the probe before evaluation. This has limitations in usage as once the probe is removed and replaced, the accuracy declines since the classifier performance is sensitive to the probe location on the arm. In this paper, we propose training a model on multiple data collection sessions to create a generalized model, utilizing incremental learning through fine tuning. Ultrasound data was acquired for 5 hand gestures within a session (without removing and putting the probe back on) and across sessions. A convolutional neural network (CNN) with 5 cascaded convolution layers was used for this study. A pre-trained CNN was fine tuned with the convolution blocks acting as a feature extractor, and the parameters of the remaining layers updated in an incremental fashion. Fine tuning was done using different session splits within a session and between multiple sessions. We found that incremental fine tuning can help enhance classification accuracy with more fine tuning sessions. After 2 fine tuning sessions for each experiment, we found an approximate 10\% increase in classification accuracy. This work demonstrates that incremental learning through fine tuning on ultrasound based hand gesture classification can be used improves accuracy while saving storage, processing power, and time. It can be expanded to generalize between multiple subjects and towards developing personalized wearable devices.
\end{abstract}

\begin{IEEEkeywords}
Ultrasound, Machine Learning with Biosignal Processing, Fine Tuning, Data Adaptation, Incremental Learning
\end{IEEEkeywords}

\section{Introduction}
% \subsection{Domain Adaptation and LLMs}
Domain adaptation is a widely used strategy with large language models (LLMs) to modify pre-trained models for improved task accuracy \cite{guan2021domain}. One approach that avoids retraining these models is fine tuning, wherein a small amount of data is used to adjust the model parameters for concluding layers of a model \cite{xu2021gradual}. This allows the model to train for significantly less time and require less computational power while maintaining similar accuracy \cite{xin2024parameter}. As particularly large models require lots of computing power and time, this is an efficient strategy \cite{church2021emerging}. Fine tuning can also be used for image classifiers, and it has been used for classifying fine art \cite{cetinic2018fine}, and medical image analysis \cite{tajbakhsh2016convolutional}.
%\subsection{Ultrasound-based hand gesture estimation}

Ultrasound data from the forearm has been shown to predict hand gestures \cite{bimbraw2022prediction}, finger angles \cite{bimbraw2023simultaneous}, and finger forces \cite{bimbraw2023estimating}. This has been used towards robotic control \cite{bimbraw2020towards}, virtual reality (VR) interfacing \cite{bimbraw2023leveraging}, and in rehabilitation use cases \cite{hettiarachchi2015new}. In our previous research, we have demonstrated the capability of our CNN to accurately classify hand gestures based on forearm ultrasound, including during online evaluation in a custom VR environment \cite{bimbraw2023leveraging}. While this system works well when the data collection and online model evaluation all occur during a singular session (without the ultrasound probe removed) we have found that the accuracy decreases significantly when the online model evaluation occurs during a separate session (with the ultrasound probe removed and replaced before online evaluation). We believe this to be caused by differences in the placement of the probe on the forearm between sessions. To resolve this, we investigated modifying our system to train on multiple data collection sessions with the ultrasound probe removed and replaced to make the model generalize over sessions. We found that this resolved the previously mentioned issue of the system losing accuracy when performing online model evaluation during a separate session, however, it took significantly longer to train the model, used more computational power, and required significantly more storage, as we used 10 times more data. To resolve this, we resorted to fine-tuning as a more efficient way to use less computational power, time and data storage.
% %Due to this, the ability to accurately determine what hand position is being made is crucial for these use cases. Ultrasound is one means that can be applied for this purpose. By recording ultrasound images of the forearm and processing them through a machine learning (ML) system, hand gestures can be accurately classified. A system can then be developed to utilize these gestures for usage as opposed to a more conventional control system.
% \subsection{Problems with the current systems}
% \subsection{How fine tuning can help}

Fine-tuning can help resolve this by allowing us to train the model on a segment of the data set and then freeze the top layers of the model (which will then act as a generalized feature extractor) only training the bottom layers on the new data. This allows us to maintain high accuracy even with isolated data sets without needing the long model training time, high data storage, and high computational needs that go along with it. Section II describes the methodology and Section III describes the experimental design. Section IV and V describe the results and future work.
% \subsection{What this paper proposes}
\section{Methods}
Forearm ultrasound data, captured using a SonoQue L5 linear wireless ultrasound probe, was streamed to a Windows 10 system with a mobile NVIDIA RTX 5000 Ada Generation graphics processor and an Intel Core i9-13950HX 24-core processor. The data collected was used to validate the effectiveness of fine-tuning compared to traditional training.
\subsection{Data Acquisition}
\begin{figure}[htbp]
\centering
\includegraphics[width=250pt]{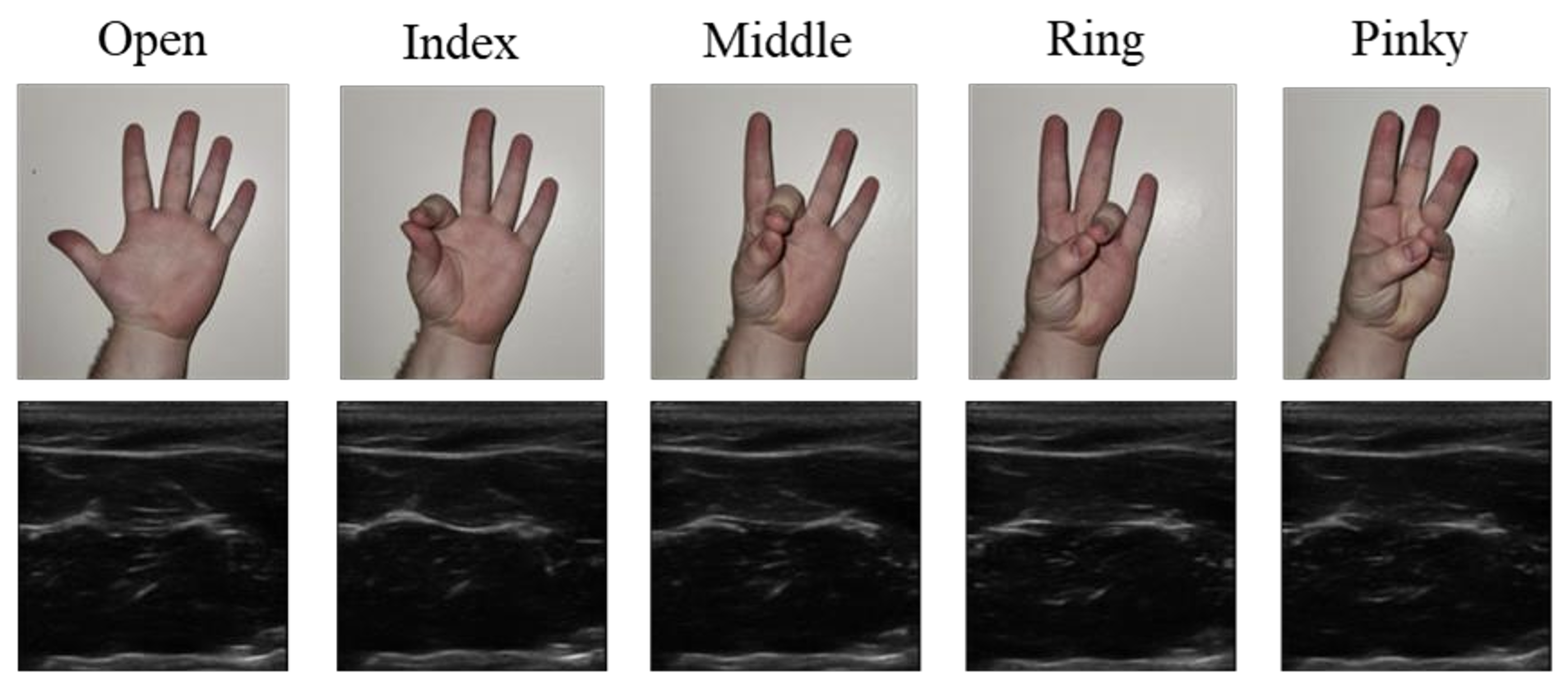} 
\caption{Hand gestures and corresponding ultrasound images}
\label{fig1}
\end{figure}

A single subject was enlisted for this research project approved under IRB-23-0634. The subject was seated at a table with their arm secured with the wireless ultrasound probe with the collection script running. The subject was instructed of the different hand positions and when to change positions in a manner similar to \cite{bimbraw2023leveraging}. A Python based data collection script helped save the forearm ultrasound images from the probe. Audio beeps were used to signify the change in hand gestures, which were labelled and organized for ease of analysis. The 5 hand gestures used for the study are shown in Figure \ref{fig1}. There were 5 rounds per data collection session with 200 images being captured of each gesture performed. This was repeated for 10 data collection sessions, in which the probe was removed from the subject, a 15-minute break was taken, and then the experiment was set back up between each session. 
\subsection{Data Preprocessing}
B-mode ultrasound data is streamed to a Windows system with a proprietary software for displaying ultrasound images. Screenshots of the image are then acquired, which are then gray-scaled to be fed to the model for training and evaluation. For this study, 640 x 640 pixel ultrasound images were used. The data was subject to different train-test splits for different experiments.
\subsection{Model description}
A CNN which has been shown to work well for hand gesture classification was used for this study, based on \cite{bimbraw2022prediction, bimbraw2023estimating}. This CNN has 5 cascaded convolution sections followed by flattening and then two dense layers which lead to the 1x5 output for predicting the different hand position classes for the open, index, middle, ring, and pinky positions based on a single frame of a forearm ultrasound image.
% \begin{figure}[htbp]
% \centering
% \includegraphics[width=255pt]{Model.png} 
% \caption{The CNN with 5 convolutional sections followed by flattening and then two dense layers which lead to the 1x4 output for predicting the finger angles for the index, middle, ring, and pinky fingers based on a single frame of a forearm ultrasound image.}
% \label{fig}
% \end{figure}
\subsection{Model training}
Adam optimizer was used for training, with sparse categorical cross entropy loss. The model was trained on the data from 5 sessions with a learning rate of 0.001. To perform the fine-tuning, the convolution sections were frozen to act as a generalized feature extractor and the learning rate was decreased to 0.000001. The model was then fine-tuned on an additional sessions, with details provided in Section III.
\section{Experimental Design}
To evaluate the performance with fine tuning, 2 experiments were performed. For both experiments, the data was split into a section for initial training, sections for fine tuning, and then a section for evaluation. The evaluation data was kept isolated to keep it independent of the training data. First, the CNN was trained normally on the initial training set and then evaluated. Then, the first round of fine tuning occurs, in which all layers but the last are frozen to act as a feature extractor. The learning rate is then reduced and the model is trained again on the first fine tuning data section before being evaluated on the evaluation set. This repeats for all fine tuning data sections. This allows for the demonstration of the accuracy increasing as incremental fine tuning occurs. 

The evaluation on the test set was done for 3 cases: For the vanilla case, the initial few sets were used for for training. For FT1, the model trained on the initial few sets was fine tuned with 1 additional set of data. For FT2, the FT1 model was successively fine tuned with another set of the data. In total, 7 sessions of data collection were acquired for the study. Each session involved 5 rounds, each recording all 5 classes for 200 images per class. This resulted in a total of 5000 images per session, with 35000 total images used for training and evaluation. 
\subsection{Intra-session experiment}
\begin{figure}[htbp]
\centering
\includegraphics[width=250pt]{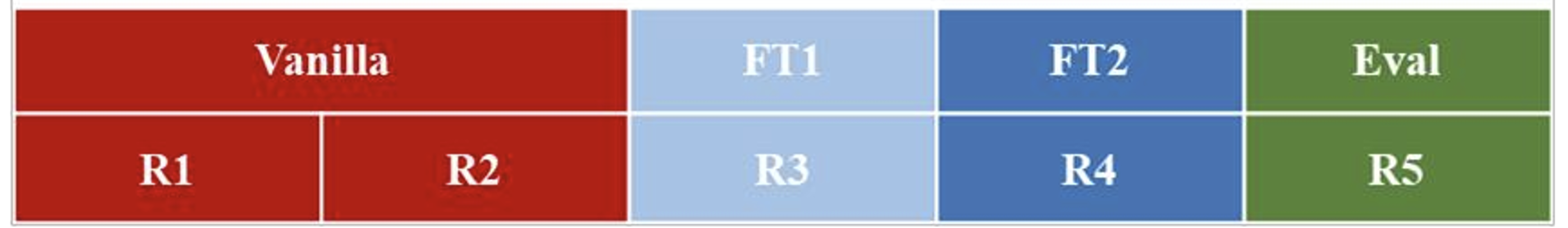} 
\caption{Data split for the intra-session experiment.}
\label{fig2}
\end{figure}

The first experiment was for testing intra-session fine tuning to verify that incremental fine tuning can perform well in an intra-session scenario. The data split is described in Figure \ref{fig2}. A single data collection session was utilized with the different rounds being used as different data sections. The initial training occurred on the first 2 rounds, then 2 sessions of incremental fine tuning occurred with the next 2 rounds, with all being evaluated on the final round.
\subsection{Inter-session experiment}
\begin{figure}[htbp]
\centering
\includegraphics[width=250pt]{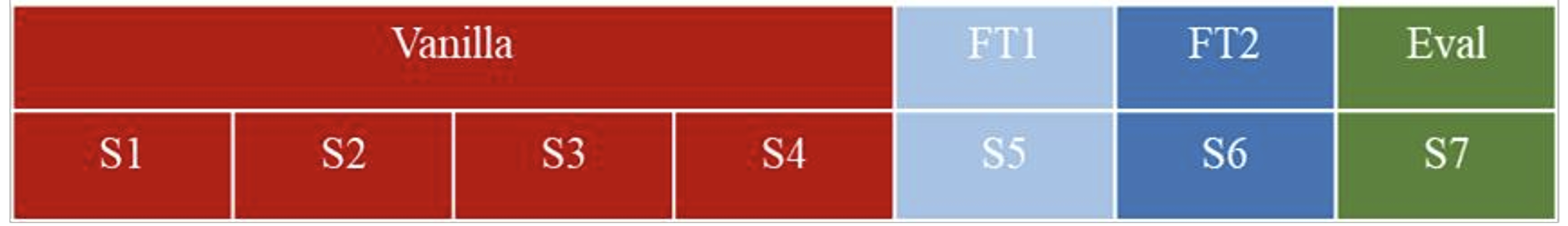} 
\caption{Data split for the inter-session experiment.}
\label{fig3}
\end{figure}

The second experiment was for testing inter-session fine tuning with the splits described in Figure \ref{fig3}. All 7 data collection sessions were utilized with the different sessions being used for the different sections. The initial training occurred on the first 4 sessions, then 2 sessions of incremental fine tuning occurred with the next 2 sessions, with all being evaluated on the final session.
\section{Results}
Comparing the accuracy of each model (the vanilla model and each fine tuned model) on the same evaluation set helped us determine if there are improvements with using this method. To further ensure that our results are consistent across different splits of the data, we performed cross-validation, repeating the experiment multiple times while randomly assigning different training, evaluation and fine tuning sets. We found that both experiments showed increases in accuracy on the evaluation set with an approximate 10\% increase after 2 incremental fine tuning sessions. Additionally, an approximate 2\% increase was observed in the second fine tuning session compared to the first fine tuning session for both experiments.
\subsection{Intra-session results}
\begin{figure}[htbp]
\centering
\includegraphics[width=250pt]{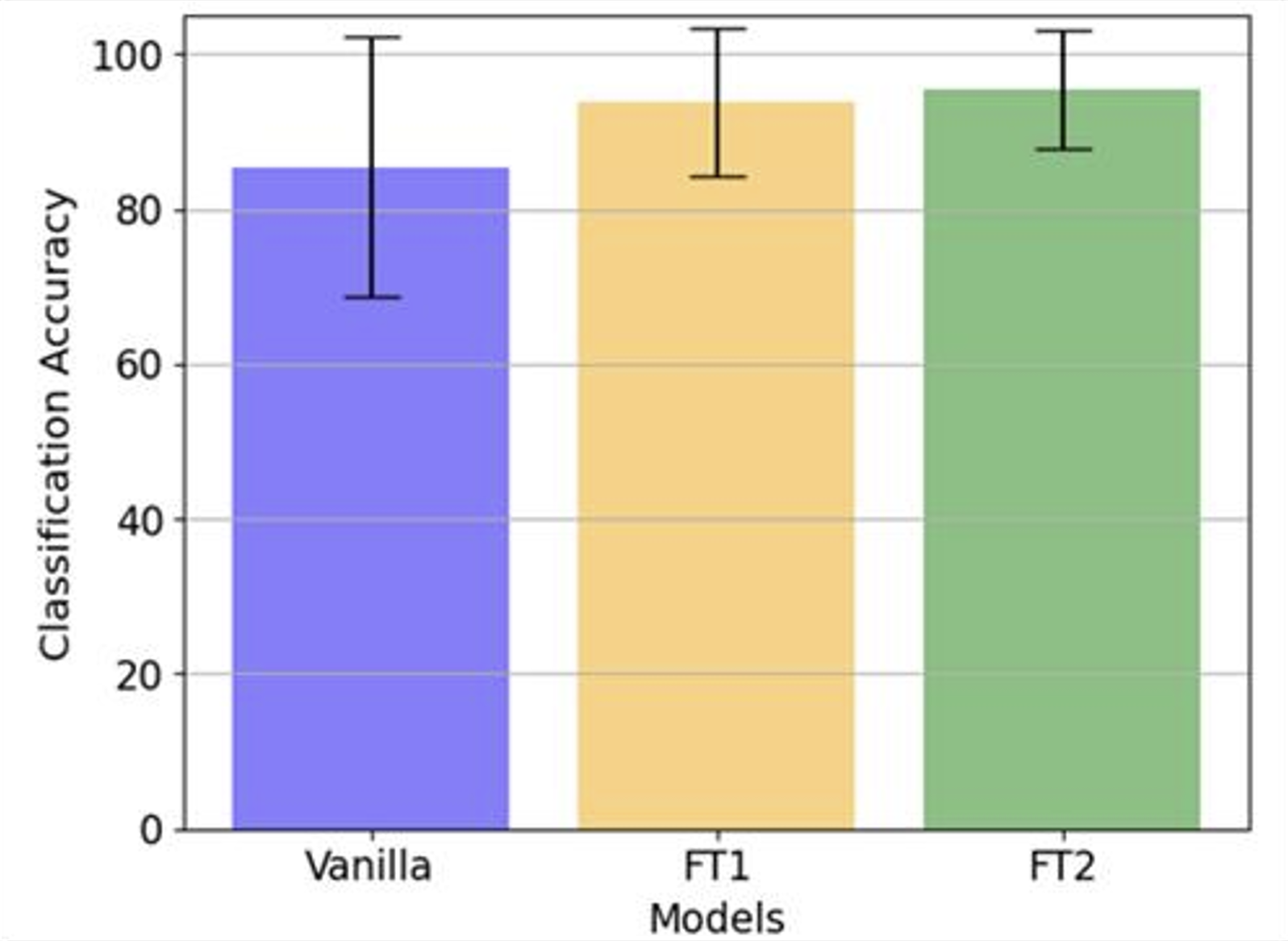} 
\caption{Averaged cross validation results for the intra-session experiment. Incremental fine tuning showed 10.1\% performance improvement compared to the vanilla case.}
\label{results1}
\end{figure}

The vanilla model had an average accuracy of 85.4\% with a standard deviation of 16.8\%. The model after 1 round of fine tuning had an average accuracy of 93.8\% and a standard deviation of 9.5\%. After the 2nd round of fine tuning, the model had an average accuracy of 95.5\% and a standard deviation of 7.6\%. That is a total of 10.1\% increase in accuracy and a 9.2\% decrease in the standard deviation from the incrementally fine tuned model over the vanilla approach. These results are shown in Figure \ref{results1}.
\subsection{Inter-session results}
\begin{figure}[htbp]
\centering
\includegraphics[width=250pt]{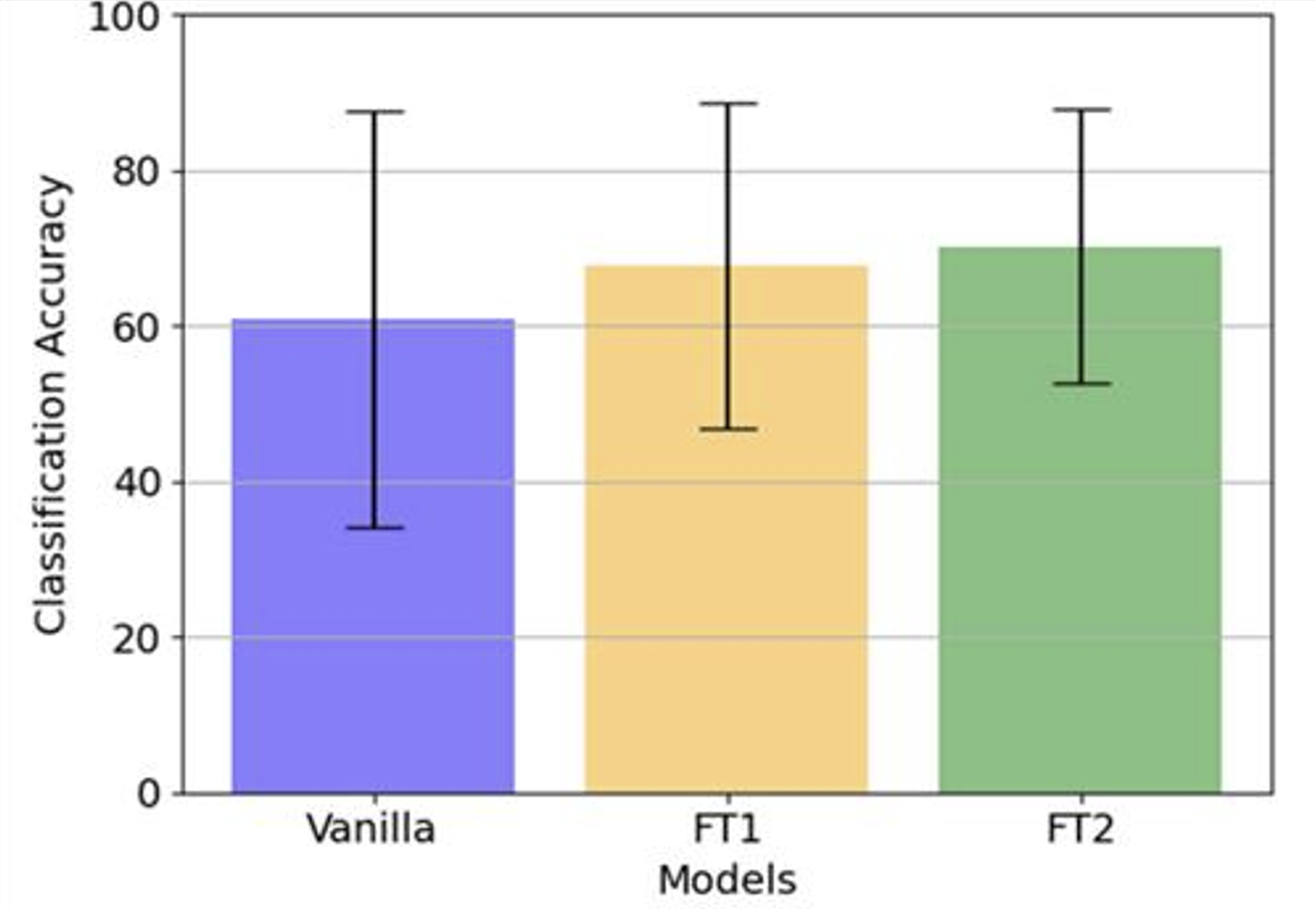} 
\caption{Averaged cross validation results for the inter-session experiment. Incremental fine tuning showed a 9.3\% performance improvement compared to the vanilla case.}
\label{results2}
\end{figure}

The vanilla model had an average accuracy of 61.0\% with a standard deviation of 26.8\%. The model after 1 round of fine tuning had an average accuracy of 67.9\% and a standard deviation of 21.0\%. After the 2nd round of fine tuning, the model had an average accuracy of 70.3\% and a standard deviation of 17.7\%. That is a total of 9.3\% increase in accuracy and a 9.1\% decrease in the standard deviation from the incrementally fine tuned model over the vanilla method. The results are shown in Figure \ref{results2}.
\section{Discussion}
These results show significant improvement of accuracy and standard deviation for the fine tuned models compared to the vanilla model over the same evaluation set for both experiments. Specifically, the accuracy increases and the standard deviation decreases after each round of incremental fine tuning. This is ideal as greater accuracy means better performance, and a smaller standard deviation means more consistent performance. Notably, for both experiments, the accuracy incrementally improves and standard deviation incrementally decreases with successive fine tuning.
\subsection{Future work}
In previous work, we were able to accurately classify hand positions from a single subject during online evaluation during that same session \cite{bimbraw2023leveraging}. In this work, we were able to improve upon that system by generalizing the model and utilizing fine-tuning so the system maintains its accuracy between sessions. In future work, this can be further expanded by generalizing the system between multiple subjects. With interesting research on miniaturizing ultrasound data acquisition \cite{frey2022wulpus} and processing \cite{bimbraw2024forearm}, this could eventually be expanded into personalized wearables, where a large model would be loaded onto a system and then gradually improve over time with incremental fine tuning.
\section{Conclusions}
This work demonstrates the validity of incremental fine tuning for forearm ultrasound based hand gesture classification. We verified that incremental fine tuning can be used to improve gesture classification performance with the probe intact during training and evaluation. Additionally, we expanded that to analyze cases where the probe is removed and then attached back for each session. We obtained improvements in classification accuracy and reduction in standard deviation for both the experiments with incremental fine tuning. This has potential towards developing personalized models, in addition to advantages such as using smaller amounts of data, computational power, and time compared to approaches without incremental fine tuning. 

% \printbibliography

\end{document}